\documentclass{ieeeaccess}
\usepackage{cite}
\usepackage{amsmath,amssymb,amsfonts}
\usepackage{algorithmic}
\usepackage{graphicx}
\usepackage{tabularx}
\usepackage{amsmath}
\usepackage{url}
\usepackage{subcaption}
\usepackage{float}
\usepackage{textcomp}
\usepackage{xcolor}
\usepackage{soul}

\renewcommand\hl[1]{#1}

\def\BibTeX{{\rm B\kern-.05em{\sc i\kern-.025em b}\kern-.08em
    T\kern-.1667em\lower.7ex\hbox{E}\kern-.125emX}}
\begin{document}
\history{Received 14 February 2025, accepted 6 March 2025, date of publication 18 March 2025, date of current version 26 March 2025.}
\doi{0.1109/ACCESS.2025.3552409}

\title{Performance Modeling of Data Storage Systems using Generative Models}
\author{\uppercase{Abdalaziz Al-Maeeni}\authorrefmark{1},
\uppercase{Aziz Temirkhanov}\authorrefmark{2}, \uppercase{Artem Ryzhikov}\authorrefmark{2}, and Mikhail Hushchyn.\authorrefmark{2}}

\address[1]{HSE University, Moscow, Myasnitskaya Ulitsa, 20, 101000, Russia (e-mail: al-maeeni@hse.ru)}
\address[2]{HSE University, Moscow, Myasnitskaya Ulitsa, 20, 101000, Russia}
\tfootnote{%%The publication was supported by the grant for research centers in the field of AI provided by the Analytical Center for the Government of the Russian Federation (ACRF) in accordance with the agreement on the provision of subsidies (identifier of the agreement 000000D730321P5Q0002) and the agreement with HSE University No. 70-2021-00139.
\hl{The article was prepared within the framework of the project “Mirror Laboratories” HSE University.}
The computation for this research was performed using the computational resources of HPC facilities at HSE University. ~\cite {Kostenetskiy_2021}}

\markboth
{Author \headeretal: Preparation of Papers for IEEE TRANSACTIONS and JOURNALS}
{Author \headeretal: Preparation of Papers for IEEE TRANSACTIONS and JOURNALS}

\corresp{Corresponding author: Mikhail Hushchyn (e-mail: mhushchyn@hse.ru).}

\begin{abstract}
High-precision systems modeling is one of the main areas of industrial data analysis. Models of systems, their digital twins, are used to predict their behavior under various conditions. In this study, we developed several models of a storage system using machine learning-based generative models to predict performance metrics such as IOPS and latency. The models achieve prediction errors ranging from 4\%–10\% for IOPS and 3\%–16\% for latency and demonstrate high correlation (up to 0.99) with observed data. By leveraging Little’s law for validation, these models provide reliable performance estimates. Our results outperform conventional regression methods, offering a vendor-agnostic approach for simulating data storage system behavior. These findings have significant applications for predictive maintenance, performance optimization, and uncertainty estimation in storage system design.

\end{abstract}


\titlepgskip=-21pt

\maketitle

\section{Introduction}
\label{sec:introduction}
\PARstart{D}{ata} storage systems (DSS) are of vital importance in the modern world. The emergence of big data has necessitated the development of storage systems with a larger capacity and lower costs to efficiently store and process the vast amounts of data generated \cite{Byron2018Using, Li2013Efficient}. Performance modeling is essential in the development of such systems.  The modeling allows engineers to analyze the system's behavior under different conditions and helps them to find the optimal design of the system. It can be used for marketing purposes to estimate the system performance under the customer's requirements. One more application is diagnostics and predictive maintenance, where model predictions are compared with actual measurements to detect failures and anomalies.

The key components of a typical DSS are controllers, fast cache memory, and storage pools. All data is stored in the pools consisting of several hard disk drives (HDDs) or solid-state drives (SSDs) united with raid schemes. The cache speeds up read and write operations with the most popular data blocks. The controllers are responsible for data management and provide computational resources to process all user input and output requests. 

An early attempt to model SSDs was described in \cite{huangPerformanceModelingAnalysis2011}. The authors argue that HDD performance models cannot be used for SSD due to certain unique characteristics, such as low latency, slow update, and expensive block-level erasure. The black-box approach to model SSD performance is suggested as it requires minimal \textit{a priori} information about the device. The researchers found that although black-box simulation does not produce high-quality predictions for HDDs, it can produce accurate ones for SSDs.

Another paper presents a black-box approach based on regression trees to model performance metrics for SSDs \cite{liBlackBoxPerformanceModeling2010}. The resulting model can accurately predict latency, bandwidth, and throughput with mean relative errors of 20 \%, 13 \%, and 6 \%, respectively.

The authors of paper \cite{kimPerformanceModelingPractical2021} propose SSDcheck, a novel SSD performance model, which is capable of extracting various internal mechanisms and predicting the latency of the next access to commodity SSDs. SSDcheck can dynamically manage the model to accurately predict latency and extract useful internal mechanisms to fully exploit SSDs. Additionally, the paper presents multiple practical cases evaluated to show significant performance improvement in various scenarios. The proposed use cases also leverage the performance model and other features to achieve an improvement of up to 130 \% in overall throughput compared to the baseline.

Machine learning techniques have proved to be useful in the task of analyzing the disks. As an example, the main approaches discussed in \cite{tarihiQuickGenerationSSD2022}  are based on machine learning. The paper uses a methodology that extracts low-cost history-aware features to train SSD performance models that predict response times to requests. The paper also utilizes space-efficient data structures such as exponentially decaying counters to track past activity with $O(1)$ memory and processing cost. Lastly, the paper uses machine learning models such as decision trees, ensemble methods, and feedforward neural networks to predict the storage hardware response time. All of these approaches are designed to make accurate SSD simulators more accessible and easier to use in real-world online scenarios.

The authors of paper \cite{zengIMRSimDiskSimulator2022} developed an open-source disk simulator called IMRSim to evaluate the performance of two different allocation strategies for interlaced magnetic recording (IMR) technology using a simulation designed in the Linux kernel space using the Device Mapper framework and offers a scalable interface for interaction and visualization of input and output (I/O) requests.

Another open-source simulator was developed in \cite{jungSimpleSSDModelingSolid2018}. The authors introduce SimpleSSD, a high-fidelity holistic SSD simulator. Aside from modeling a complete storage stack, this simulator decouples SSD parallelism from other flash firmware modules, which helps to achieve a better simulation structure. SimpleSSD processes I/O requests through layers of a flash memory controller. The requests are then serviced by the parallelism allocation layer, which abstracts the physical layout of interconnection buses and flash disks. Although aware of internal parallelism and intrinsic flash latency, the simulator can capture close interactions among the firmware, controller, and architecture. 

The idea of SimpleSSD has been further developed in \cite{gouk2018amber}. SimpleSSD 2.0, namely Amber, was introduced to run in a full-system environment. Being an SSD simulation framework, Amber models embedded CPU cores, DRAMs, and various flash technologies by emulating data transfers. Amber applies parallelism-aware readahead and partial data update schemes to mimic the characteristics of real disks. 

Research that used different Flash Translation Layer (FTL) schemes was presented in \cite{5283998}. FlashSim, an SSD simulator, has been developed to evaluate the performance of storage systems. The authors analyzed the energy consumption in different FTL schemes to find realistic workload traces and validated FlashSim against real commercial SSDs and detected behavioral similarities.

Although simulating SSD devices might have seemed an exciting research topic, older storage systems have also received the scientific attention they deserve. One of the papers presents a series of HDD simulations to investigate the stiffness of the interface between the head and the disk. In \cite{jayson2002shock}, the authors develop a finite-element HDD model using a commercially available three-dimensional modeling package. The results of the numerical analysis provided insight on the characteristics of each type of shock applied to an HDD model.

Although used extensively in modern times, SSDs are not designed perfectly. There is a certain performance-durability trade-off in their design. Disks of this type require garbage collection (GC), which hinders I/O performance, while the SSD block can only be erased a finite number of times. There is research that proposes an analytical model that optimizes any GC algorithm \cite{li2013stochastic}. The authors propose the randomized greedy algorithm, a novel GC algorithm, which can effectively balance the above-mentioned trade-off. 

In the search for the perfect architecture of a modern SSD through design space exploration (DSE), researchers have come across the challenge of fast and accurate performance estimation. In \cite{kimFastPerformanceEstimation2020}, the authors suggest two ways to tackle this challenge: scheduling the task graph and estimation based on neural networks (NN). The proposed NN regression model is based on training data, which consists of hardware configurations and performance. The paper concludes that the NN-based method is faster than scheduling-based estimation and is preferred for scalable DSE. 

The evaluation of SSD performance has also been investigated in \cite{kangFrameworkEstimatingExecution2017}. SSDs are built using an integrated circuit as memory, so manufacturers are interested in I/O traces of applications running on their disks. This paper proposes a framework for accurately estimating I/O trace execution time on SSDs.

Malladi et al. \cite{malladi2017flashstoragesim} introduces FlashStorageSim, an SSD architecture performance model validated with an enterprise SSD. It models the SSD controller, flash devices, and host interface (e.g., SATA, PCIe, DDR), allowing for the exploration of various SSD use cases. The model includes mechanisms to improve simulation speed, achieving over 7X reduction in simulation time. Validation against a real SSD shows that FlashStorageSim accurately captures performance trends. The fast-forwarding (FF) and cycle granularity adjustment (CGA) mechanisms further enhance simulation speed. The methodology involves developing FlashStorageSim in C++ to model SSD components and interfaces and incorporating mechanisms to reduce simulation time. The model is validated against a real server SSD using IO benchmarks and trace collection to compare performance metrics like response time and bandwidth. Speed optimization is assessed through fast-forwarding and cycle granularity adjustment, showing simulation speed improvements

The work done by Lebre et al. \cite{7152491} presents advancements in the simulation of storage resources for distributed computing systems. The main results include the successful extension of SimGrid with comprehensive storage simulation capacities, introducing a versatile API that handles storage components and their contents. The authors characterized the performance of various types of disks, deriving accurate models that account for sharing and contention effects. These models were validated through extensive experiments on the Grid'5000 testbed, demonstrating their practicality and accuracy. The paper also highlighted several potential use cases for the enhanced SimGrid toolkit, such as optimizing data placement and replication policies in MapReduce applications, improving storage management in grids and clouds, and accurately simulating hierarchical storage systems in data centers. The proposed extension thus enables more accurate and comprehensive simulation studies, facilitating better design and management of storage infrastructure in large-scale distributed computing environments.

Louis et al. \cite{7431390} extends the CloudSim toolkit \cite{neves2011cloudsim} to include a scalable module for modeling and simulating energy-aware storage in cloud systems. The main results demonstrate that CloudSimDisk effectively simulates the energy consumption of storage systems and validates against analytical models. CloudSimDisk introduces HDD models and power models, supports energy-aware simulations, and includes features for disk array management and transaction processing. Simulations using CloudSimDisk align closely with analytical predictions, confirming its accuracy and efficacy.

The paper \cite {lee2021learned} by Lee et al. introduces an accurate data-driven model for SSD firmware validation. By directly executing the firmware on both a simulator and the target device, the model collects performance profiles to predict SSD performance. Using a deep neural network (DNN), the model achieves a 3.1\% error rate, significantly lower than the 18.9\% error from existing simulators. This approach reduces the need for extensive testing on prototype boards, improving efficiency and accuracy in SSD development. This work is the first to apply a data-driven method for SSD performance prediction, effectively bridging the gap between simulation and real-world performance.

\cite{wan2008simulation} Wan et al. develop a simulation model to predict the cumulative distribution function (CDF) of I/O request response time in a RAID 0 system. The methodology involves a bottom-up approach, starting with a single disk drive modeled as an M/G/1 queue. This model is then extended to a RAID 0 system using a split-merge queueing network. The model's accuracy is validated by comparing simulation results of I/O request response times with measurements from actual devices, showing close agreement. Speed optimizations include fast-forwarding and cycle granularity adjustments, improving simulation efficiency.

This work \cite{lebrecht2011analytical} presents a fork-join queueing simulation model of RAID systems with zoned disk drives, specifically focusing on RAID levels 01 and 5. Starting with an M/G/1 queue-based simulation for a single disk, the model is extended to RAID using a fork-join network. The simulator accepts heterogeneous I/O workloads and produces the I/O request response time distribution, incorporating request-reordering optimizations and validating against device measurements. The simulation model accurately predicts I/O request response times for RAID 01 and 5 systems, showing agreement with actual device measurements. It highlights the impact of request size, workload burstiness, and request-reordering optimizations on performance. The model simulates RAID systems under various conditions, providing insights for improving analytical models and optimizing RAID configurations.

Other work in the context of storage system modeling includes various approaches. Low-level simulators for magnetic tapes \cite{johnson1998performance, sandsta1999low}, hard drives \cite{bucy2003disksim}, and SSDs \cite{kim2009flashsim} use detailed discrete-event simulations. DiskSim \cite{bucy2003disksim}, which models block-level operations, faces scalability issues. SIMCAN \cite{nunez2012simcan} extends this with layered simulations of disks, volumes, and file systems, which can lead to configuration challenges. In HPC simulations, scalability is often prioritized over accuracy, with random I/O times used to improve performance \cite{carothers2002ross}.

MapReduce simulators \cite{kolberg2013mrsg, wang2009simulation} typically use simplistic disk models, focusing on the logic of file systems rather than their performance. This paper extends SimGrid with versatile and accurate storage simulation capabilities, leveraging its expertise in network modeling \cite{velho2013validity} and introducing an API that addresses sharing and contention effects across different infrastructures.

All of these solutions have a range of limitations. The first one is that they are focused mostly on the modeling of one single device such as HDD or SSD disks. The second is that most of them are based on simulation of physical processes and the firmware stack inside the devices, which require more resources for development. However, one of the main disadvantages is that the models do not take into account vendor specifics of the devices due to the trade secrets. As a result, the errors in performance simulation are quite high for practical use. Finally, HDDs and SSDs are part of DSS with unique architectures and software that influence the device performance and are not taken into account in the solutions described above. 

In this work, we present a data-driven approach for DSS modeling that does not have these limitations. All vendor specifics, architecture features, and software impact are learned directly from real system performance measurements. We present the performance modeling and analysis of a data storage cache, SSD, and HDD pools under random and sequential data loads. We provide novel data sets of IOPS and latency measurements for the system's components for various configurations and load parameters. We consider two different approaches to learn distributions of the performance values, based on parametric and nonparametric generative models. The inclusion of generative models allows for a more comprehensive simulation of system behavior. More precisely, the learned distributions encompass every potential performance value corresponding to the provided data load parameters. By employing these values, we can determine the means and standard deviations of IOPS and latency, and examine the correlations between them. In contrast to conventional regression models, which are limited to predicting only mean values. 

We analyze the quality of the model predictions and compare them with a naive baseline algorithm. We also demonstrate a physics-inspired method for checking the reliability of the model predictions. The data sets provided in this work can be used for benchmarking regression algorithms, conditional generative models, and uncertainty estimation methods in machine learning.

Additionally, our approach comprehensively models the entire storage system, encompassing individual disks of various types, the complete array, and the cache layer, with an emphasis on delivering an integrated and holistic solution. This methodology ensures that every component, from the smallest disk to the most complex array configuration, is accounted for in our model. Moreover, our approach is designed to be universally applicable, transcending vendor-specific constraints, thereby ensuring compatibility with a wide array of hardware platforms. This vendor-agnostic nature of our model enhances its versatility, making it suitable for diverse hardware environments and enabling seamless integration across different storage infrastructures.

\hl{The article is organized as follows. In Section 2, we define the task of predicting IOPS and latency for storage components under varying workloads and configurations. Section 3 describes the generation of datasets using tools like Perf, covering parameters such as load type, block size, RAID schemes, and job configurations. In Section 4, we introduce CatBoost and Normalizing Flows (NF) for learning performance distributions, alongside a kNN baseline. Section 5 outlines evaluation metrics to assess prediction accuracy and distributional fidelity. Section 6 presents model performance, showing CatBoost’s lower prediction errors (4–10\% for IOPS, 3–16\% for latency) and NF’s ability to capture complex variability, validated by Little’s law. Section 7 analyzes model behavior, including conservatism in extrapolation regimes, and highlights practical implications for DSS simulation. Finally, Section 8 summarizes findings, advocates for generative models in DSS performance modeling, and proposes future work on uncertainty estimation and explainability.}

% \subsection{Equations}
% Number equations consecutively with equation numbers in parentheses flush
% with the right margin, as in \eqref{eq}. To make your equations more
% compact, you may use the solidus (~/~), the exp function, or appropriate
% exponents. Use parentheses to avoid ambiguities in denominators. Punctuate
% equations when they are part of a sentence, as in
% \begin{equation}E=mc^2.\label{eq}\end{equation}

% The following 2 equations are used to test 
% your LaTeX compiler's math output. Equation (2) is your LaTeX compiler' output. Equation (3) is an image of what (2) should look like.
% Please make sure that your equation (2) matches (3) in terms of symbols and characters' font style (Ex: italic/roman).

% \begin{align*} \frac{47i+89jk\times 10rym \pm 2npz }{(6XYZ\pi Ku) Aoq \sum _{i=1}^{r} Q(t)} {\int\limits_0^\infty \! f(g)\mathrm{d}x}  \sqrt[3]{\frac{abcdelqh^2}{ (svw) \cos^3\theta }} . \tag{2}\end{align*}

% $\hskip-7pt$\includegraphics[scale=0.52]{equation3.png}

% Be sure that the symbols in your equation have been defined before the
% equation appears or immediately following. Italicize symbols ($T$ might refer
% to temperature, but T is the unit tesla). Refer to ``\eqref{eq},'' not ``Eq. \eqref{eq}''
% or ``equation \eqref{eq},'' except at the beginning of a sentence: ``Equation \eqref{eq}
% is $\ldots$ .''

\section{Problem statement}
\begin{table}
\caption{\textbf{Input and output features of the simulation models}}
% \label{table}
% \small
% \begin{tabular*}{17.5pc}{@{}|p{30pt}|p{112pt}<{\raggedright}|p{30pt}<{\raggedright}|@{}}
% \begin{tabular*}{17.5pc}{@{}|p{30pt}|p{112pt}|p{30pt}|@{}}
\begin{tabular}{|p{25pt}|p{150pt}|p{40pt}|}
% {\linewidth}{@{}p{2cm} @{}p{8.5cm}<{\raggedright} X@{}}
\hline
Model & 
Inputs & 
Outputs \\
\hline
Storage pool & 
Load type, IO type, read fraction, block size, number of jobs, queue depth, number of disks, number of data, and parity blocks in RAID & 
IOPS, Latency \\
\hline
Storage cache & 
Load type, IO type, read fraction, block size, number of jobs, queue depth & 
IOPS, Latency \\
\hline
\end{tabular}
\label{tab1}
\end{table}

In this study, we consider the simulation of HDD and SSD storage pools and cache, which are the key parts of data storage systems. The goal of this work is to predict the performance of these components for a given configuration and data load parameters. We describe the performance by the number of input and output operations per second (IOPS) and the average of their latencies. The data load is parameterized by: load type, IO type, read fraction, block size, number of jobs, and queue depth for each of these jobs. There are two load types we consider in the work: random and sequential. Each data load consists of a mix of read and write operations. The IO type indicates whether we consider read or write operations in our simulations. The read fraction is the ratio of the number of read operations to the total number of all operations in the data load. Each operation processes one data block of a given size. The data loads are generated by several jobs with a given queue depth.

HDD and SSD storage pools also have configuration parameters. They are the total number of disks in a pool and the software RAID scheme, which is described by the number of data and parity blocks. In this work, the cache has only one configuration and we do not describe it by any parameter. The lists of input and output features in our simulation models are shown in \textbf{Table~\ref{tab1}}.

Let us denote the vector of input features as $x$, and the vector of outputs as $y$. Also suppose we have $n$ pairs of measurements $\{x_i, y_i\}_{i=1}^{n}$ of output vectors $y_i$ for the given inputs $x_i$. These pairs are obtained by generating various data loads on a real data storage system and measuring the performance of its pool or cache in terms of IOPS and latency. The goal of this work is to estimate the conditional probability distribution $p(y|x)$ for the storage component using machine learning. Then, we predict the performance $\hat{y}_j$ for unknown inputs $x_j$ by sampling from the learned distribution:

\begin{equation}
\hat{y}_j \sim p(y|x_j).
\end{equation}

The advantage of this data-driven approach is that we do not need to understand and simulate all the physical processes inside the storage components. The model only needs a sample of measurements. Machine learning algorithms estimate all physical dependencies from the data. We can use this model to explore and visualize the performance dependencies from the configuration and data load parameters, to predict the average performance and its deviations for new conditions, and to analyze many other properties.

\section{Data collection}\label{sec:data}
\begin{table}[t]

\caption{\textbf{Data load parameters and their value ranges for the storage cache data set}}
%\label{table}
% \small
% \begin{tabular*}{17.5pc}{@{}|p{60pt}|p{125pt}<{\raggedright}|@{}}
%\begin{tabular*}{17.5pc}{@{}|p{30pt}|p{112pt}|p{30pt}|@{}}
% \begin{tabularx}{\textwidth}{@{}p{7cm} X@{}}
\begin{tabular}{|p{50pt}|p{110pt}|}
\hline
Parameter & 
Value range \\
\hline
Load type & random \\
%IO type & read, write \\
Block size & 4, 8, 16, 32, 64, 128, 256 KB \\
Read fraction & 0 - 100\% \\
Number of jobs & 1 - 64 \\
Queue depth & 1 - 16 \\
\hline
\end{tabular}
\label{tab:data-cache}
\end{table}

\begin{table}[t]
\caption{\textbf{Data load parameters and their value ranges for the storage SSD pool data set; random data loads}}
%\label{table}
\small
\begin{tabular}{|p{60pt}|p{120pt}|}
% {@{}|p{60pt}
% |p{125pt}<{\raggedright}
% |@{}
% }
%\begin{tabular*}{17.5pc}{@{}|p{30pt}|p{112pt}|p{30pt}|@{}}
\hline
Parameter & 
Value range \\
\hline
Load type & random \\
%IO type & read, write \\
Block size & 4, 8, 16, 32, 64 KB \\
Read fraction & 0 - 100\% \\
Number of jobs & 1 - 32 \\
Queue depth & 1 - 32 \\
\hline
RAID (K+M) & 1+1, 2+1, 2+2, 4+1, 4+2, 8+2 \\
Number of disks & K+2M, 24, +3 values in between  \\
\hline
\end{tabular}
\label{tab:data-ssd-rnd}
\end{table}

\begin{table}[t]
\caption{\textbf{Data load parameters and their value ranges for the storage SSD pool data set; for sequential data loads}}
%\label{table}
% \small
% \begin{tabularx}{\linewidth}{@{}p{7cm} X@{}}
%\begin{tabular*}{17.5pc}{@{}|p{30pt}|p{112pt}|p{30pt}|@{}}
\begin{tabular}{|p{60pt}|p{120pt}|}
\hline
Parameter & 
Value range \\
\hline
Load type & sequential \\
%IO type & read, write \\
Block size & 128, 256, 512, 1024 KB \\
Read fraction & 0\% \\
Number of jobs & 1 - 20 \\
Queue depth & 1 - 32 \\
\hline
RAID (K+M) & 1+1, 2+1, 2+2, 4+1, 4+2, 8+2 \\
Number of disks & K+2M, 24, +3 values in between  \\
\hline
\end{tabular}
\label{tab:data-ssd-seq}
\end{table}

\begin{table}[t]
\caption{\textbf{Data load parameters and their value ranges for the storage HDD pool data set; for sequential data loads}}
%\label{table}
% \small
% \begin{tabularx}{\linewidth}{@{}p{7cm} X@{}}
%\begin{tabular*}{17.5pc}{@{}|p{30pt}|p{112pt}|p{30pt}|@{}}
\begin{tabular}{|p{60pt}|p{120pt}|}
\hline
Parameter & 
Value range \\
\hline
Load type & sequential \\
%IO type & read, write \\
Block size & 128, 256, 512, 1024 KB \\
Read fraction & 0, 100\% \\
Number of jobs & 1 - 20 \\
Queue depth & 1 - 32 \\
\hline
RAID (K+M) & 1+1, 2+1, 2+2, 4+1, 4+2, 8+2 \\
Number of disks & K+2M, 24, +3 values in between  \\
\hline
\end{tabular}
\label{tab:data-hdd-seq}
\end{table}

For this study, we collected four data sets for the storage pools and cache under different data loads. The first one is for the cache under the random data loads. To collect it, we generated 510 different data loads using a performance analysis tool Perf~\cite{perf} in Linux. The tool generated a stream of data blocks from a client to the system for read and write operations and collected statistics about the processing of these blocks. \textbf{Table~\ref{tab:data-cache}} provides the list of the load parameters and the ranges of their values. We quasi-randomly selected these values for each data load using the Sobol sequence~\cite{sobol1967distribution}. This method covers the parameter space more uniformly than random or grid sampling. Each load lasted 60 seconds, yielding every second IOPS and average latency for read and write operations separately, giving us two separate results with respect to read and write operations. In this case, IOPS means how many data blocks from the client were read or written to the cache per one second. As a result, we collected $510\times60\times2=61200$ pairs of measurements $(x_i, y_i)$. 

Similarly, we collected data sets for the SSD pool under random loads. The list of all parameters and their value ranges is in \textbf{Table~\ref{tab:data-ssd-rnd}}. In this case, we also used the pool configuration parameters in addition to the data load ones. The first is a software RAID scheme, which is defined by the number of data (K) and parity (M) blocks. The second is the total number of disks in the pool. The minimum amount of disk in the pool can be determined by applying the K+2M formula listed in the table. E.g. for RAID schema 4+2, the minimum number of disks required is 8, and the maximum is 24 for any configuration. We include minimum, maximum, and 3 values in between, like 12, 16, and 19 disks in that example. We generated 512 different loads in total for various pool configurations with a duration of 120 seconds each. We measured IOPS and average latency for read and write operations separately every second. In the results, this data set contains $512\times120\times2 = 122820$ pairs of measurements $(x_i, y_i)$. 

We also collected two data sets for SSD and HDD storage pools under sequential data loads. We used the same procedure described above. The main difference is that only pure loads with read fractions of 0 and 100 \% are valuable. One more difference is that larger block sizes are used for sequential loads. \textbf{Table~\ref{tab:data-ssd-seq}} and \textbf{Table~\ref{tab:data-hdd-seq}} show the full list of the data load and configuration parameters and their value ranges. We used the Sobol sequence to generate 512 different loads in total for various pool configurations with a duration of 120 seconds each. We measured IOPS and average latency for read and write operations every second. In the results, each data set contains $512\times120=61440$ pairs of measurements $(x_i, y_i)$. 
As a result, we collected 4 distinct data sets, with different amounts of rows. Additionally, we engendered one feature by multiplying the depth and number of jobs and splitting the raid schema variable into two variables. Instead of string feature K+M therefore obtaining K and M. Thus, for SSD and HDD pools modeling we use 10 features, and for cache - only 7.

\hl{The value ranges were derived from the system specification. Due to the sheer size of the parameter ranges, and the amount of all possible combinations, it is more sensible to use only a subsample of the vast parameter space. The Sobol sequence was chosen because it ensures more uniform coverage of the parameter space compared to random selection. It is important because it ensures to sample from rare or critical regions, and that samples are equally distanced to avoid clustering in high-dimensional space.  Thus,  leading towards better generalization ability of the model.}
\section{Models} 
\label{sec:models}

\subsection{CatBoost model}

\begin{figure*}
\centerline{\includegraphics[width=1\textwidth]{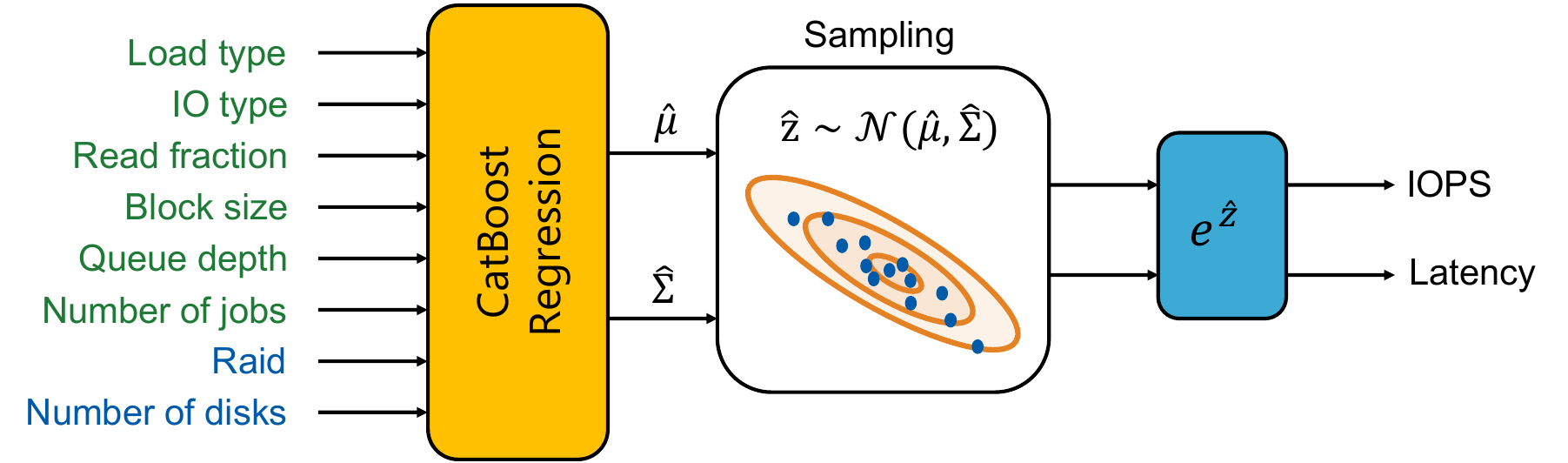}}
\caption{The CatBoost-based model for performance predictions of the storage pools and cache for the given values of data load and configuration (for the pools only) parameters.}
\label{fig:catboost}
\end{figure*}

The first model we use in this study is a parametric generative model based on the CatBoost~\cite{catboost2018} regression model. Relations between IOPS and latencies within the same data load are defined by Little's law~\cite{10.2307/167570}:

\begin{equation}
\begin{split}
Q \times J & = IOPS_{read} \times Latency_{read} + \\
 & IOPS_{write} \times Latency_{write},
\end{split}
\end{equation}

where $Q$ is the queue depth and $J$ is the number of jobs. While the fraction of read operations is fixed during each data load, Little's law allows us to suppose the following relation for read or write operations:

\begin{equation}
\log IOPS \sim -\log Latency.
\end{equation}

All measurements of IOPS and latencies are stochastic. We approximate the distribution of  their logarithm values by conditional 2D normal distributions:

\begin{equation}
\hat{z}_i = \log \hat{y}_i,
\end{equation}
\begin{equation}
\hat{z}_i \sim \mathcal{N}(\hat{\mu}(x_i), \hat{\Sigma}(x_i)),
\end{equation}

where $\hat{y}_i$ is a vector of predictions for IOPS and latency; the mean $\hat{\mu}(x_i)$ and the covariance matrix $\hat{\Sigma}(x_i)$ depend on input vector $x_i$ of data load and configuration parameters, and are predicted by the CatBoost regression model. \textbf{Figure~\ref{fig:catboost}} shows the model.

As described in the previous section, a sample contains $2m$ read and write data loads with $k$ measurements $\{x_i, y_i\}_{i=0}^{k}$ in each for just read or write operations. The total number of measurements is $n=2mk$. We calculate the mean vectors $\mu_j$ and the covariance matrices $\Sigma_j$ for each of these data loads. In addition, we use Cholesky decomposition~\cite{10.5555/248979} for the matrices $\Sigma_j^{-1}=L_jL_j^T$. It is used to ensure that the predicted covariance matrices will be positive semi-definite. We fit the CatBoost regression model with the MutliRMSE loss function defined as:

\begin{equation}
\begin{split}
L^2 & = \frac{1}{2m}\sum_{j=1}^{2m} (\hat{\mu}(x_j)-\mu_j)^2 + \frac{1}{2m}\sum_{j=1}^{2m} (\hat{L}(x_j)-L_j)^2,
\end{split}
\label{eq:mutlirmse}
\end{equation}

where $\hat{\Sigma}^{-1}(x_i)=\hat{L}(x_j)\hat{L}(x_j)^T$. 

The regression model is implemented with \\ \textit{CatBoostRegressor}, with specific parameters and an extensive hyperparameter tuning process to optimize its performance. The \textit{CatBoostRegressor} is configured to use the \textit{MultiRMSE} loss function, which is designed for multivariate regression tasks, aiming to minimize the root mean square error across multiple output variables. The same metric, \textit{MultiRMSE}, is employed to evaluate the model's performance during training, ensuring consistency in optimization and evaluation.

The model is fitted on the mean vectors $\mu_j$ and the covariance matrices $\Sigma_j$ with a maximum of 5000 boosting iterations, with a learning rate of 0.1 to control the step size during each iteration, thus preventing overfitting by reducing the influence of each individual tree. Early stopping is not employed in this setup, allowing the model to run for all specified iterations. To ensure the reproducibility of the results, a random seed of 42 is used. The model is also configured to return the best iteration based on the evaluation metric, enhancing its final performance. Additionally, training on data with constant labels is permitted by setting the \textit{allow\_const\_label} parameter to \textit{True}.

To identify the optimal hyperparameters, a grid search is conducted over a defined parameter space. The depth of the trees varies across values of 2, 4, 6, and 8, while the learning rate is tested at 0.01, 0.05, and 0.1. This systematic search allows for a comprehensive exploration of the hyperparameter space to find the best combination that improves the predictive accuracy of the model.

The model training involves fitting the \textit{CatBoostRegressor} on the training dataset and evaluating it on a validation dataset. During this process, the grid search parameters are utilized, providing a mechanism to fine-tune the model’s performance. Upon completion of the training process, the best model, as determined by the evaluation metric, is saved to a specified path, ensuring that the optimal model configuration is preserved for future use. See \footnote{https://github.com/HSE-LAMBDA/digital-twin} for a concrete implementation.

\subsection{Normalizing flow}

\begin{figure*}
\centerline{\includegraphics[width=\textwidth]{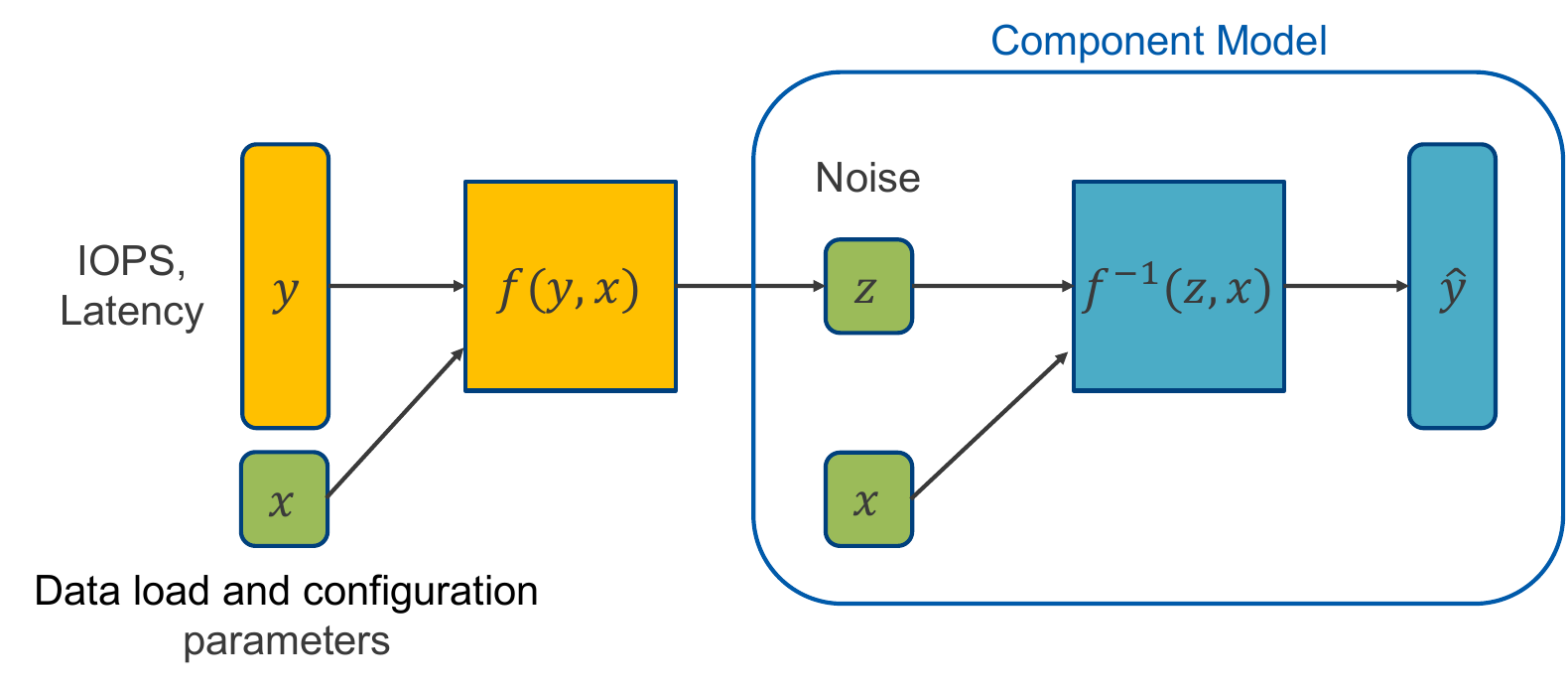}}
\caption{The normalizing flow model for performance predictions of the storage pools and cache for the given values of data load and configuration (for the pools only) parameters}
\label{fig:nf}
\end{figure*}

The second model we apply in this work is based on Normalizing Flows (NF)~\cite{JMLR:v22:19-1028}. Consider a sample with $n$ measurements $\{x_i, y_i\}_{i=1}^{i=n}$ for various data loads and configurations of a data storage component. Let us also define a random variable $z$, which is sampled from the standard normal distribution $q(z) = \mathcal{N}(0, I)$. The NF model learns an invertible transformation between the vector of measured IOPS and latency $y_i$ into the random variable $z_i$ with the given $x_i$:

\begin{equation}
z_i = f(y_i, x_i).
\end{equation}

The change of variable theorem determines the relation between the estimated performance $\hat{p}(y_i|x_i)$ and $q(z_i)$ distributions:

\begin{equation}
\hat{p}(y_i|x_i) = q(f(y_i, x_i)) \left| \det \frac{\partial f(y_i, x_i)}{\partial y_i} \right|.
\end{equation}

In this study, we consider the Real NVP normalizing flow model~\cite{realnvp}, where the function $f(y_i, x_i)$ is designed using a chain of neural networks. The model is fitted by optimizing the log-likelihood loss function:

\begin{equation}
L = \frac{1}{n} \sum_{i=1}^{n} \log \hat{p}(y_i|x_i) \to \max_{f}.
\end{equation}

Performance value predictions are made as $\hat{y}_i=f^{-1}(z_i, x_i)$, where $z_i$ is a randomly sampled vector from the $q(z)$ distribution. An illustration of the NF model is shown in {\bf Figure~\ref{fig:nf}}. In this case, $\hat{y}_i$ is sampled from the learned distribution $\hat{p}(y_i|x_i)$ of the IOPS and latency with the given data load and configuration parameter values.

In our experiments, we use sixteen Real NVP transformations. Each transformation (RealNVP layer) contains two neural networks for scale and translation transformation. These NN topologically are 2-linear fully-connected neural networks, with {\it tanh} activation, and 10 hidden neurons.  We fit the model using the {\it Adam} optimizer. It is trained for 80 epochs with a batch size of 200 and a learning rate $\eta = 10^{-2}$.
After the training, such a model can transform a pure sample from Gaussian distribution to our target distribution by applying named transformations. Since the model was trained by minimizing the negative log-likelihood loss function it tends to make wider prediction, e.i. overestimating the variance of the distribution.

\subsection{kNN model}

Consider a new vector $x^{*}$ of inputs for which we need to predict IOPS and latency values. As previously, assume that our train sample contains $2m$ read and write data loads with $k$ measurements in each. The total number of observations is $n=2mk$ $\{x_i, y_i\}_{i=0}^{i=n}$. All these data loads are described by $2m$ unique input vectors $U=\{u_i\}_{j=1}^{j=2m}$. Then, for the new input $x^{*}$ the k Nearest Neighbors (kNN) algorithm searches for the nearest vector $u^{*}$:

\begin{equation}
u^{*} = \arg \min_{u_j \in U} d(x^{*}, u_j),
\end{equation}

where $d(x, z)$ is the distance between two input vectors. In this work, we used the Euclidean distance. The predictions are made as follows. The model takes for the predictions all $k$ observations $y_i$ for which $x_i=u^{*}$:

\begin{equation}
\hat{y}^{*} = \{y_i | x_i = u^{*}\}.
\end{equation}

In other words, the model finds the closest known data load to the new one and takes its measured IOPS and latencies as predictions. We use this approach as a baseline in the study. This helps to estimate whether the CatBoost and the NF models based on machine learning algorithms provide better prediction results than the naive approach based on the kNN algorithm.

\subsection{Model selection}

In this work, we compare parametric (CatBoost) and non-parametric (NF) generative models for performance modeling. CatBoost was chosen due to its robustness in handling tabular data and its MultiRMSE loss function, which effectively estimates mean vectors and covariance matrices in multivariate regression tasks. On the other hand, the Normalizing Flow model complements CatBoost by learning non-parametric distributions, enabling it to handle complex variability in performance metrics. The combined use of these models allows us to balance precise predictions with the flexibility to model uncertainty, aligning well with the requirements of data storage system performance modeling. They are based on different ideas and have their own advantages and limitations. The parametric model assumes the target is described by a normal distribution with parameters $\hat{\mu}(x_i)$ and $\hat{\Sigma}(x_i)$ for the given input $x_i$. The CatBoost regression model learns these parameters separately using the MultiRMSE loss function. Prediction errors for the $\hat{\mu}(x_i)$ and $\hat{\Sigma}(x_i)$ are independent. For example, a large error for the mean vector does not affect the accuracy of the covariance matrix prediction and vice versa. We use the CatBoost because it is one of the best regression models for tabular data. 

The nonparametric model (NF) learns the distribution of the observed data without assumptions about its shape. It can learn any distribution, not only the normal one. The model optimizes the log-likelihood loss function, and all distribution parameters are learned dependently. For example, if the center of a learned distribution is shifted, its width can also be wider than the real one to provide the minimum of the loss function. So, the model optimizes the whole distribution. 

The CatBoost and NF models fit data distributions differently. And one model can be better than another, depending on the quality metric. Further in this work, we will compare these models using a range of various metrics.

\section{Quality metrics} \label{sec:metrics}

\begin{figure*}
\centerline{\includegraphics[width=1.\textwidth]{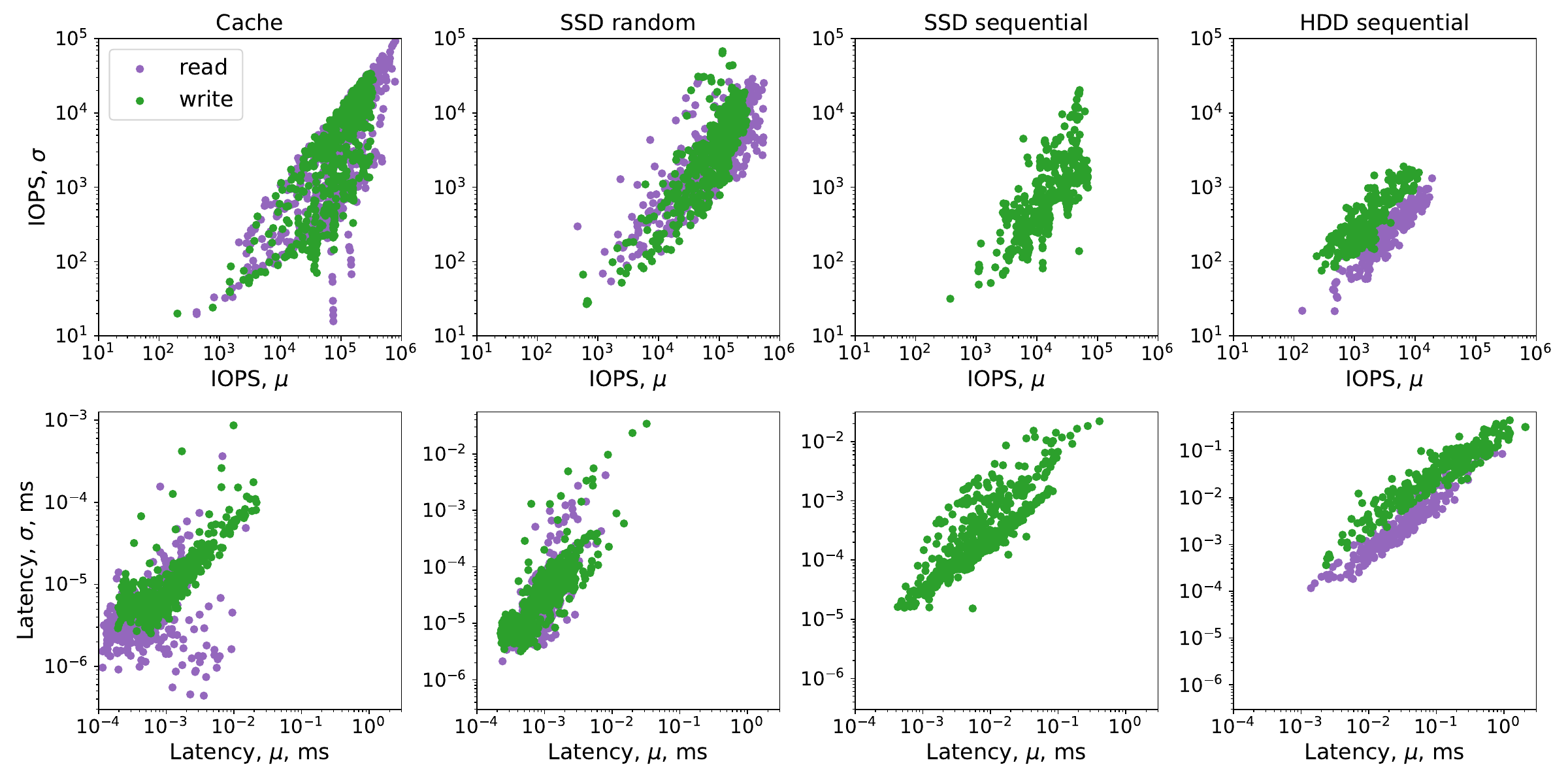}}
\caption{Means ($\mu$) and standard deviations ($\sigma$) of the measured IOPS and latencies for cache, SSD, and HDD pools under random and sequential data loads; each point corresponds to one data load and one component configuration}
\label{fig:iops_lat_mu_sigma}
\end{figure*}

Consider the measurements $\{x_i, y_i\}_{i=0}^{k}$ for just read or write operations in one single data load. According to the data sets description, this sample contains 120 (60 for cache) measurements, where all $x_i$ have the same input feature values, and $y_i = (\text{IOPS}_i, \text{Latency}_i)^T$ is a vector of measured performance values. Also, suppose that we have predictions $\{x_i, \hat{y}_i\}_{i=0}^{k}$ from one of the models for the same $x_i$. The goal is to estimate the discrepancy between the distributions of $y_i$ and $\hat{y}_i$.

The first quality metric we use is the Percentage Error of Mean (PEM), which describes the ability of the models to predict mean values of IOPS and latency for each data load:

\begin{equation}
\text{PEM} = \left|{\frac{\hat{\mu} - \mu}{\mu}}\right|\times 100\%,
\end{equation}

\begin{equation}
\mu = \frac{1}{k}\sum_{i=1}^{k} y_i.
\end{equation}

Similarly, we use the Percentage Error of Standard deviation (PES), which describes the ability of the models to predict standard deviations of IOPS and latency values for each data load:

\begin{equation}
\text{PES} = \left|{\frac{\hat{\sigma} - \sigma}{\sigma}}\right|\times 100\%,
\end{equation}

\begin{equation}
\sigma = \sqrt{\frac{1}{k-1}\sum_{i=1}^{k} (y_i-\mu)^2}.
\end{equation}

The means ($\mu$) and standard deviations ($\sigma$) of the measured IOPS and latencies for cache, SSD, and HDD pools under random and sequential data loads are shown in \textbf{Figure~\ref{fig:iops_lat_mu_sigma}}. 

We also use two additional metrics to measure the distances between the distributions of $y_i$ and $\hat{y}_i$. The first is the Fréchet distance (FD)~\cite{DOWSON1982450}, which is widely used to estimate the quality of generative models~\cite{10.5555/3295222.3295408}. We suppose that the vectors $y_i$ and $\hat{y}_i$ have 2D Gaussian distributions $\mathcal{N}(\mu, \Sigma)$ and $\mathcal{N}(\hat{\mu}, \hat{\Sigma})$ respectively. Then, the FD is defined as:

\begin{equation}
\begin{split}
\text{FD} & = \| \mu -\hat{\mu} \|^{2}_{2} + tr(\Sigma + \hat{\Sigma} - 2(\Sigma\hat{\Sigma})^{1/2} ).
\end{split}
\end{equation}

One more metric is the Maximum Mean Discrepancy (MMD)~\cite{JMLR:v13:gretton12a}, which is defined as:

\begin{equation}
\begin{split}
\text{MMD} = \frac{1}{k^2}\sum_{i=1}^{k}\sum_{j=1}^{k} K(y_i, y_j) + \frac{1}{k^2}\sum_{i=1}^{k}\sum_{j=1}^{k} K(\hat{y}_i, \hat{y}_j) - \\ \frac{2}{k^2}\sum_{i=1}^{k}\sum_{j=1}^{k} K(y_i, \hat{y}_j), 
\end{split}
\end{equation}

where $K(u, v) = C(\sigma) exp(-\frac{|u - v|^2}{2\sigma^2})$ is the Radial Basis Function (RBF) with normalization constant $C(\sigma)$ and $\sigma$ equals the median distance between the vectors in the combined sample $y_i$ and $\hat{y}_i$.

In our study, IOPS and latency have different scales which affect the calculation of the FD and MMD. To solve this problem, we fit the Standard Scaler~\cite{scikit-learn} on the real observations $\{y_i\}_{i=1}^{k}$ and apply it to transform the values of $y_i$ and $\hat{y}_i$. Then, the scaled vectors are used for the FD and MMD estimation.

We split data loads in each data set into train and test samples. Test one contains 100 randomly chosen load configuration with all their measurements (meaning 120 or 60 measurements for each fixed parameters' set, and for read and write settings) $\{x_i, y_i\}$, therefore preventing leakage to the test set.  We calculate the quality metrics for each model on each data load. Then, we use the bootstrap technique to find their average and standard deviation over all loads in the test sample. We use random shuffle with replacements for resampling and 100 steps of bootstrapping followed by mean and standard deviation calculations for the resulting arrays.

\section{Results}

\begin{table}[t]
\caption{\textbf{Quality metrics for the storage cache under random data load}}
% \label{table}
% \small
% \begin{tabularx}{\linewidth}{X@{}X@{}X@{}X@{}}
%\begin{tabular*}{17.5pc}{@{}|p{30pt}|p{112pt}|p{30pt}|@{}}
\begin{tabular}{p{53pt}p{48pt}p{48pt}p{48pt}}
\hline
Metric & 
kNN & 
CatBoost &
NF \\
\hline
PEM (IOPS), \%  & 26$\pm$2 & 6.3$\pm$0.4 & 4.1$\pm$0.4 \\
PEM (Lat), \%   & 18$\pm$1   & 4.9$\pm$0.3 & 2.9$\pm$0.2 \\
PES (IOPS), \%  & 35$\pm$2 & 23.9$\pm$0.8 & 412$\pm$23 \\
PES (Lat), \%   & 30$\pm$1   & 24.1$\pm$0.9 & 299$\pm$29 \\
FD              & 6655$\pm$822 & 445$\pm$57   & 365$\pm$55 \\
MMD             & 1.61$\pm$0.01 & 1.34$\pm$0.02 & 0.59$\pm$0.02 \\
\hline
\end{tabular}
\label{tab:metrics_cache_rnd}
\end{table}

\begin{table}[t]
\caption{\textbf{Quality metrics for the SSD pool under random data load}}
% \label{table}
% \small
% \begin{tabularx}{\linewidth}{X@{}X@{}X@{}X@{}}
\begin{tabular}{p{53pt}p{48pt}p{48pt}p{48pt}}
\hline
Metric & 
kNN & 
CatBoost &
NF \\
\hline
PEM (IOPS), \%  & 38$\pm$3        & 8.9$\pm$0.6     & 10.1$\pm$0.8 \\
PEM (Lat), \%   & 19.2$\pm$0.9    & 7.4$\pm$0.4     & 8.8$\pm$0.6 \\
PES (IOPS), \%  & 52$\pm$3        & 22$\pm$0.9      & 183$\pm$14 \\
PES (Lat), \%   & 40$\pm$2        & 25$\pm$1.1      & 135$\pm$9 \\
FD              & 1647$\pm$211    & 140$\pm$19      & 161$\pm$16 \\
MMD             & 1.38$\pm$0.02   & 1.05$\pm$0.02   & 0.83$\pm$0.02 \\
\hline
\end{tabular}
\label{tab:metrics_ssd_rnd}
\end{table}

\begin{table}[t]
\caption{\textbf{Quality metrics for the SSD pool under sequential data load}}
% \label{table}
% \small
% \begin{tabularx}{\linewidth}{X@{}X@{}X@{}X@{}}
%\begin{tabular*}{17.5pc}{@{}|p{30pt}|p{112pt}|p{30pt}|@{}}
\begin{tabular}{p{53pt}p{48pt}p{48pt}p{48pt}}
\hline
Metric & 
kNN & 
CatBoost &
NF \\
\hline
PEM (IOPS), \%  & 31$\pm$3          & 10.2$\pm$0.9     & 9.9$\pm$1 \\
PEM (Lat), \%   & 42$\pm$3          & 10.7$\pm$0.7     & 8.1$\pm$0.7 \\
PES (IOPS), \%  & 90$\pm$12       & 37$\pm$5        & 134$\pm$16 \\
PES (Lat), \%   & 101$\pm$16      & 42$\pm$5      & 131$\pm$17 \\
FD              & 1692$\pm$241      & 110$\pm$18      & 89$\pm$19 \\
MMD             & 1.23$\pm$0.03     & 1.01$\pm$0.03   & 0.69$\pm$0.03 \\
\hline
\end{tabular}
\label{tab:metrics_ssd_seq}
\end{table}

\begin{table}[t]
\caption{\textbf{Quality metrics for the HDD pool under sequential data load}}
% \label{table}
% \small
% \begin{tabularx}{\linewidth}{X@{}X@{}X@{}X@{}}
%\begin{tabular*}{17.5pc}{@{}|p{30pt}|p{112pt}|p{30pt}|@{}}
\begin{tabular}{p{53pt}p{48pt}p{48pt}p{48pt}}
\hline
Metric & 
kNN & 
CatBoost &
NF \\
\hline
PEM (IOPS), \%  & 27$\pm$2   & 10.6$\pm$0.9     & 11.4$\pm$0.9 \\
PEM (Lat), \%   & 49$\pm$4   & 16$\pm$2         & 18$\pm$2 \\
PES (IOPS), \%  & 33$\pm$3     & 18$\pm$1         & 43$\pm$4 \\
PES (Lat), \%   & 60$\pm$8     & 23$\pm$2         & 62$\pm$7 \\
FD              & 96$\pm$19  & 5.0$\pm$0.6      & 6.7$\pm$0.9 \\
MMD             & 0.81$\pm$0.04   & 0.33$\pm$0.02    & 0.23$\pm$0.02 \\
\hline
\end{tabular}
\label{tab:metrics_hdd_seq}
\end{table}

\begin{figure*}
\centerline{\includegraphics[width=1.\textwidth]{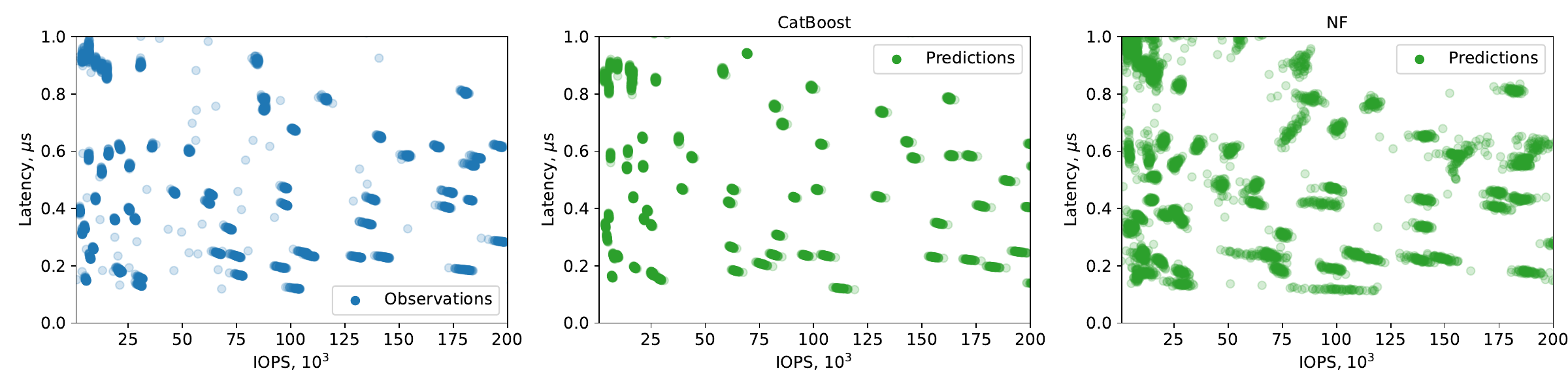}}
\caption{Example of real observations and predictions for read data loads on the cache; Each cloud corresponds to one data load.}
\label{fig:ex_cache_rnd2}
\end{figure*}

\begin{figure*}
\centerline{\includegraphics[width=1.0\textwidth]{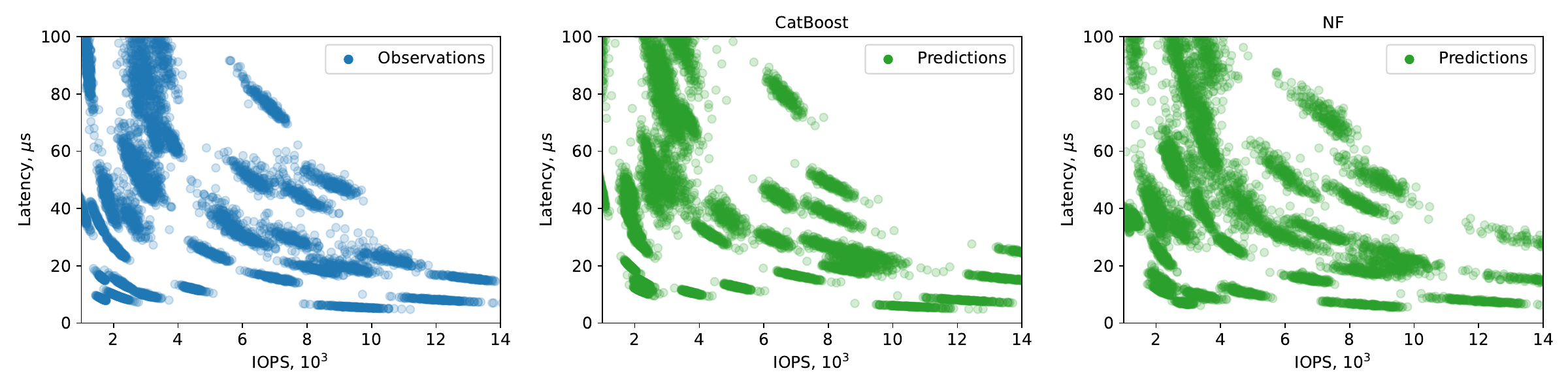}}
\caption{Example of real observations and predictions for read data loads on the HDD pool; Each cloud corresponds to one data load.}
\label{fig:ex_hdd_seq2}
\end{figure*}

\begin{figure*}
\centerline{\includegraphics[width=1.\textwidth]{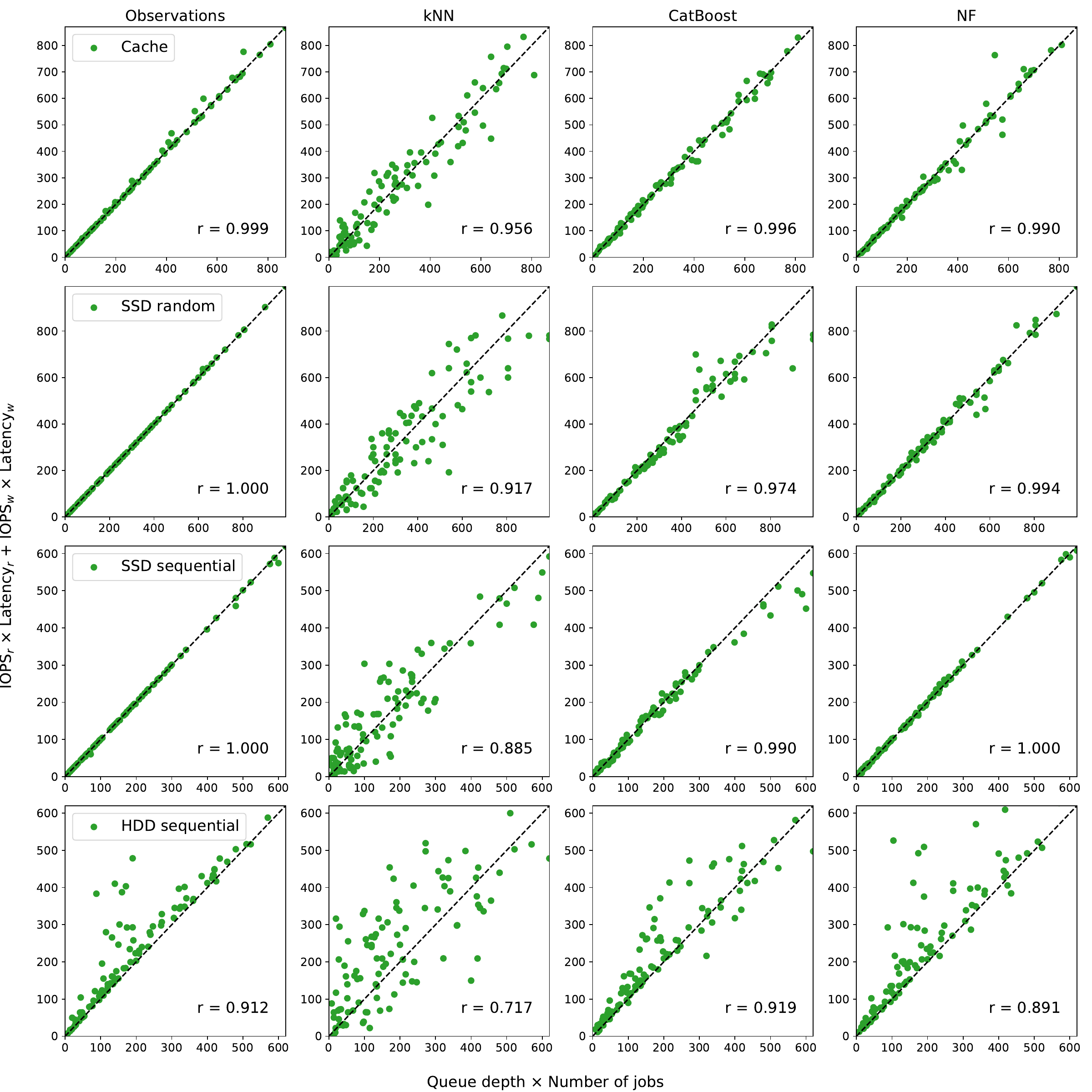}}
\caption{Little's law verification for observed and predicted IOPS and latency; each point corresponds to one data load and one configuration of a component}
\label{fig:little-law}
\end{figure*}

\begin{figure*}[h!]
\centerline{\includegraphics[width=1.0\textwidth]{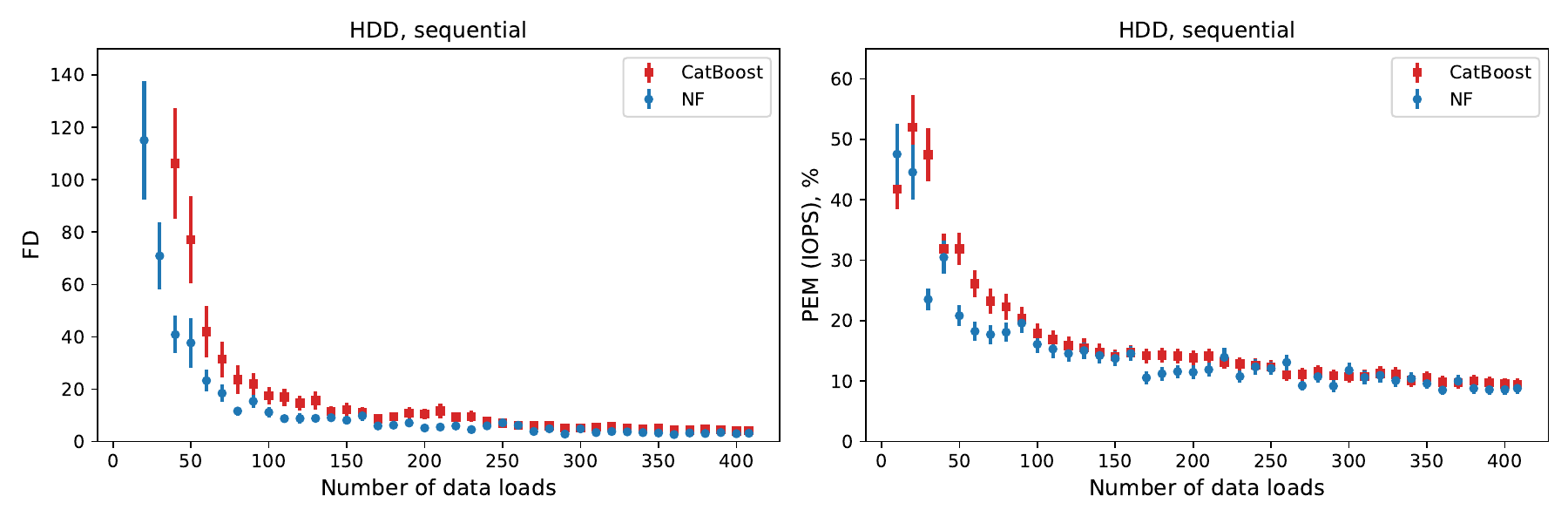}}
\caption{Dependency of prediction quality metrics from the number of data loads used in the models training for the HDD pool under sequential data load.}
\label{fig:ex_hdd_seq}
\end{figure*}

\begin{figure*}[h!]
\centerline{\includegraphics[width=1.\textwidth]{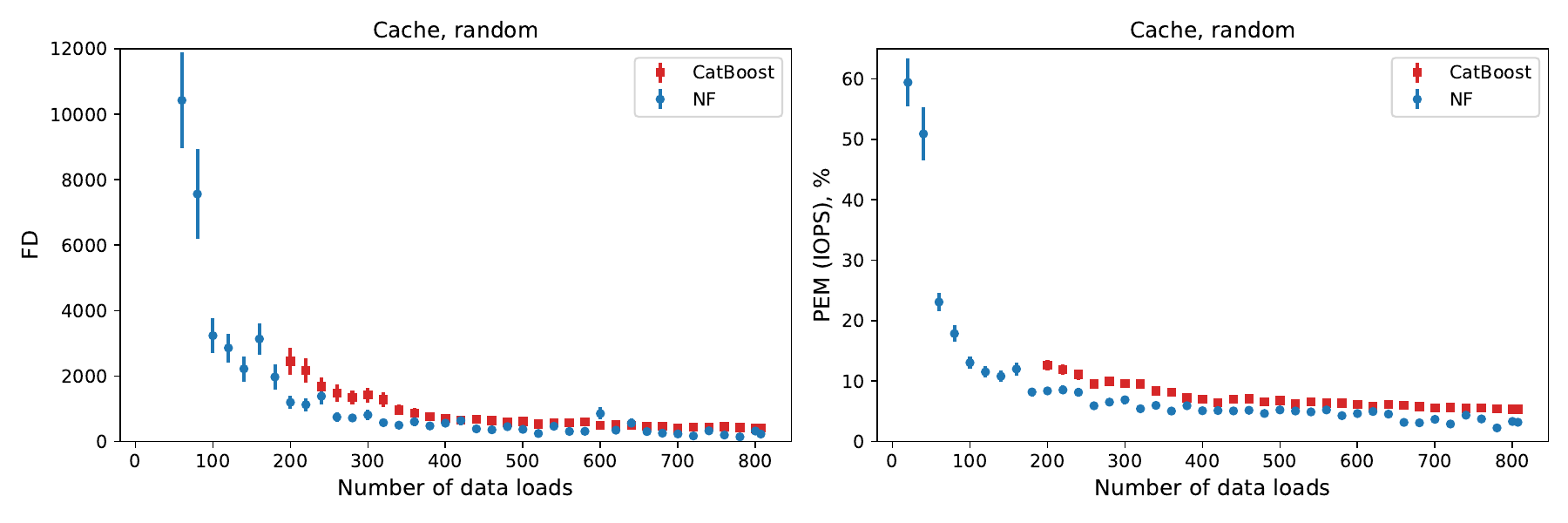}}
\caption{Dependency of prediction quality metrics from the number of data loads used in the models training for the cache under random data load.}
\label{fig:ex_cache_rnd}
\end{figure*}

\begin{figure*}[h!]
    \centering
    \begin{subfigure}{.48\textwidth}
        \includegraphics[width=\linewidth]{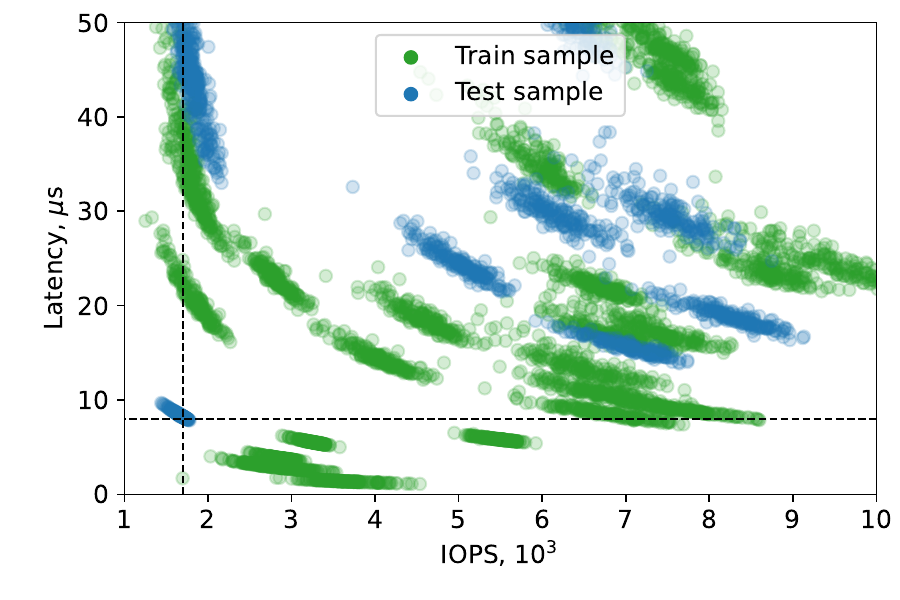}
        \caption{Example of real observations in the train and test samples for the HDD pool.}
        \label{fig:anomaly_hdd}
    \end{subfigure}%
    \hfill
    \begin{subfigure}{.48\textwidth}
        \includegraphics[width=\linewidth]{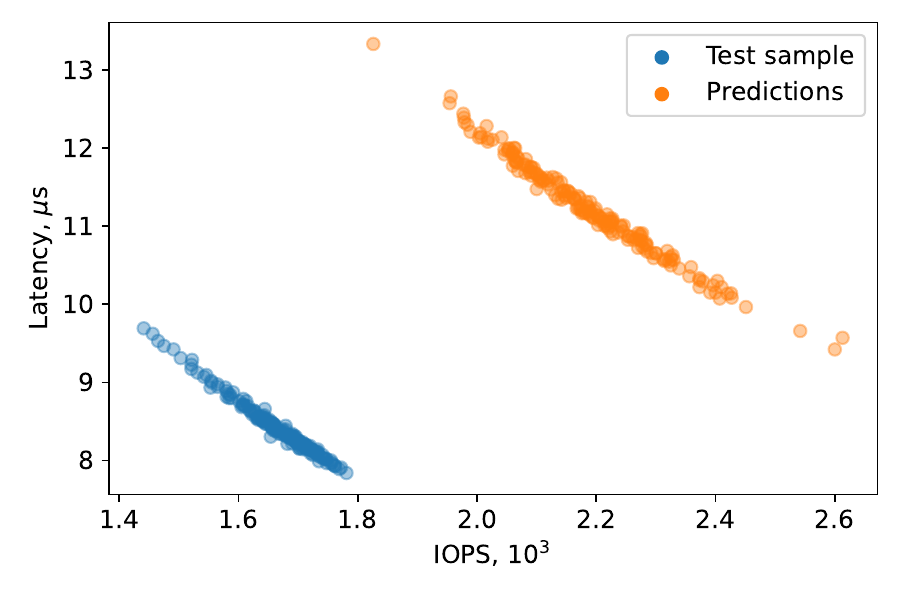}
        \caption{Example of real observations and CatBoost predictions for one data load on the HDD pool.}
        \label{fig:anomaly_hdd_exam}
    \end{subfigure}\\
    \begin{subfigure}{.48\textwidth}
        \includegraphics[width=\linewidth]{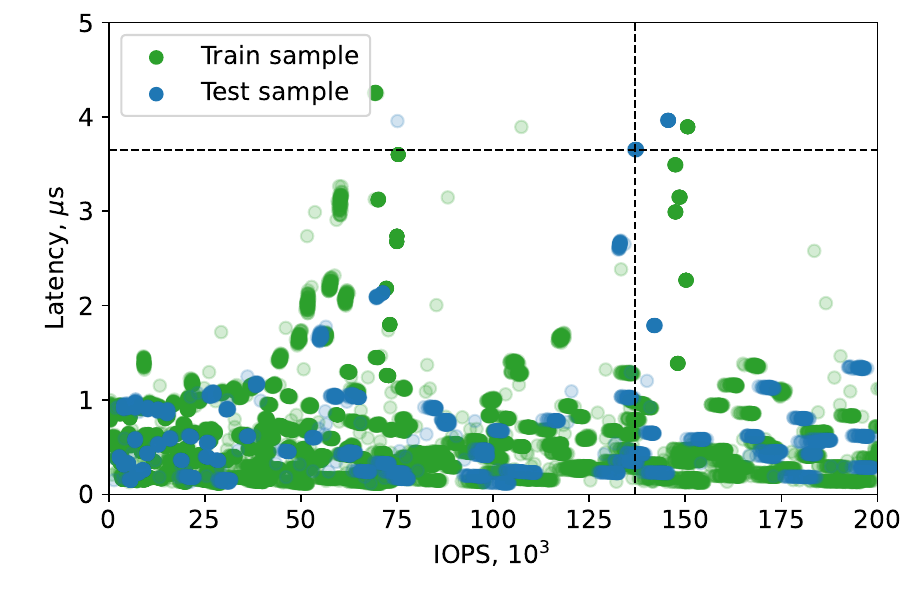}
        \caption{Example of real observations for read data loads in the train and test samples for the cache.}
        \label{fig:anomaly_cache}
    \end{subfigure}%
    \hfill
    \begin{subfigure}{.48\textwidth}
        \includegraphics[width=\linewidth]{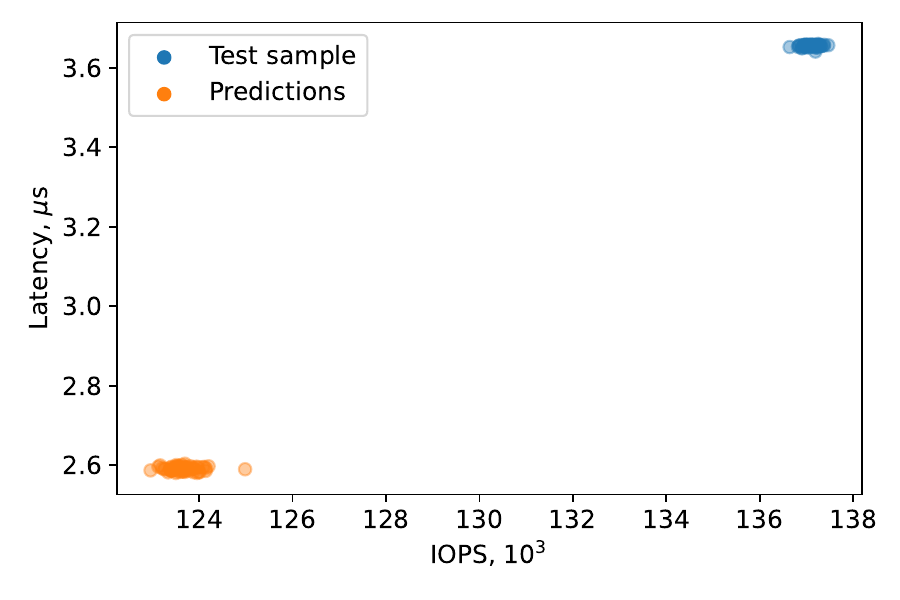}
        \caption{Example of real observations and CatBoost predictions for one data load on the cache.}
        \label{fig:anomaly_cache_exam}
    \end{subfigure}
    \caption{Anomaly Detection Results}
\end{figure*}

In this section, we consider the experiments that we have conducted to estimate the quality of the models\footnote{\url{https://github.com/HSE-LAMBDA/digital-twin}}. We fitted all models described above on the same train samples, made predictions on the same test samples, and calculated quality metrics from the previous section. The metrics values for the cache, SSD, and HDD pools are presented in \textbf{Tables \ref{tab:metrics_cache_rnd}, \ref{tab:metrics_ssd_rnd}, \ref{tab:metrics_ssd_seq}, and \ref{tab:metrics_hdd_seq}}. The models in our study learn the conditional distributions of IOPS and latency. Their predictions are samples from these distributions. Each metric highlights different quality aspects of the prediction, which we detail below.

While the NF model demonstrates superior performance for mean predictions, it tends to overestimate standard deviations, as reflected in the higher Percentage Error of Standard deviation (PES) values in Tables 6–9. This limitation arises from the model’s tendency to predict wider distributions, particularly in sparsely sampled regions of the input parameter space. Improvements could be achieved by expanding the training data set to include a wider variety of configurations and more experiments with different architectures, which could yield a better estimate of the standard deviation.

Table~\ref{tab:metrics_cache_rnd} provides the metrics values for the kNN, CatBoost, and NF models for the cache under random data loads. \textbf{Figure~\ref{fig:ex_cache_rnd2}} shows an example of the predictions for several data loads from the test sample. The PEM and PES metrics in the table show the estimation quality of the means and standard deviations, respectively, for IOPS and latency in each data load. The observed means and standard deviations are presented in Figure~\ref{fig:iops_lat_mu_sigma}. The NF model demonstrates the best PEM values, where the prediction errors are about 4.1 \% and 2.9 \% for IOPS and latency, respectively. It is about 6 times smaller than the 25 \% and 17.8 \% errors for the kNN model. Similarly, CatBoost demonstrates the smallest PES values of 23.9 \% and 24.1 \% for IOPS and latency, respectively. The NF model has the most significant errors of 412 \% and 299 \% for IOPS and latency. Generally, the results show that the standard deviation estimation is a challenging task for all models in our study. But the NF predicts the widest distributions, as demonstrated in Figure~\ref{fig:ex_cache_rnd}.

FD compares the mean values and the covariance matrices of the predictions and real observations. The metric value for the CatBoost and NF is approximately 15 and 18 times smaller than for the kNN, respectively but still quite large. This can be explained by the following two reasons. The first is the standard scaling transformation that we applied before the metric calculation. It was fitted on real observations and applied to the predictions as we described in the previous section. FD estimates the distance between two distributions on a scale, where the observed IOPS and latencies have zero means and standard deviations of 1. The second reason is that the cache is the fastest component in the data storage system. The standard deviations of its IOPS and latency measurements are relatively small compared to the mean values for other components of the system, as shown in Figure~\ref{fig:iops_lat_mu_sigma}. So, the calculated distances between the observed and predicted distributions are larger for the cache than for other components. The MMD metric compares two distributions by calculating distances between pairs of points. The distances are divided by the median values, making the metric more robust to different scales of the distributions. The results show that the NF model has the best MMD value.

Table~\ref{tab:metrics_ssd_rnd} presents the metrics values for the kNN, CatBoost, and NF models for the SSD pool of the system under random data loads. The results demonstrate that the CatBoost model is the best in terms of the FD, PEM, and PES metrics, and NF is the best in terms of the MMD metric. The CatBoost model predicts the mean of IOPS and latency values with errors of 8.9 \% and 7.4 \% respectively, compared with 38 \% and 19.2 \% for the kNN. The standard deviations of IOPS and latency are estimated with errors of 22 \% and 25 \% for CatBoost and 52 \% and 40 \% for the kNN model. Similarly to the cache, the NF model predicts wider distributions for SSD pools under random data loads as it is reflected in PES values. However, the model estimates the IOPS and latency distributions better in terms of the MMD metric. We explain this behavior as follows. NF learns wider distributions than other models. Despite these distributions having worse PES values, they cover the observations better than other models.

Similarly, Table~\ref{tab:metrics_ssd_seq} shows the metrics values for the SSD pool under sequential data loads. The results demonstrate that the CatBoost model is the best in terms of the FD, and PES metrics, and the NF is the best in terms of the MMD and PEM metrics. The best predictions of the mean of IOPS and latency values have errors of 10.2\%, 10.7\%, and 9.9\%, 8.1\% respectively for CatBoost and NF, compared to 31\% and 42\% for the kNN. The standard deviations of IOPS and latency are estimated with errors of 37\% and 42\% for CatBoost and 90\% and 101\% for the kNN model. Similarly to the cache, the NF model predicts wider distributions for SSD pools under sequential data loads as it is reflected by PES values. However, the model estimates the distributions better in terms of the MMD metric.

Finally, Table~\ref{tab:metrics_hdd_seq} provides the metrics values for the HDD pool of the system under sequential data loads. \textbf{Figure~\ref{fig:ex_hdd_seq2}} shows an example of the predictions for several data loads from the test sample. The results demonstrate that the CatBoost model is the best in terms of the FD, PEM, and PES metrics, and the NF is the best in terms of the MMD metric. The best predictions of the mean of IOPS and latency values have errors of 10.6\% and 16.0\% respectively for CatBoost, compared with 27\% and 49\% for kNN. The standard deviations of IOPS and latency are estimated with errors of 18\% and 23\% for CatBoost and 33\% and 60\% for the kNN model. The NF model predicts the widest distributions, as reflected in the PES values. However, it better estimates the distributions in terms of the MMD metric.

Generally, the results demonstrate that the NF and CatBoost significantly outperform the kNN models on all data sets. The CatBoost model has better metrics values on all data sets. The NF shows the worst results for the standard deviation estimation, but the best values of the MMD metric.

To verify the reliability of the models, we conducted an additional experiment. Relations between IOPS and latencies within the same data load are defined by Little's law~\cite{10.2307/167570}:

\begin{equation}
\begin{split}
Q \times J & = IOPS_{read} \times Latency_{read} + IOPS_{write} \times Latency_{write},
\end{split}
\label{eq:little}
\end{equation}

where $Q$ is the queue depth and $J$ is the number of jobs. For each data load in test samples, we calculated the right part of this equation and compared it with the left part estimated from the load parameters. The results are provided in \textbf{Figure~\ref{fig:little-law}}. The plots demonstrate that Little's law is satisfied for the real observations in all data sets used in this study, as well as for the predictions of the CatBoost and NF models. They learn the dependency in \textbf{Equation~\ref{eq:little}} directly from the data. \textbf{Table~\ref{tab:pearson}} provides Pearson's correlation coefficients between the measurements in Figure~\ref{fig:little-law}. It shows that the coefficients for the real observations and predictions of CatBoost and NF models are similar, and support the reliability of the models. The kNN demonstrates larger differences from Little's law than other models.

Figure~\ref{fig:little-law} also shows that CatBoost and NF predictions for several data loads deviate significantly from the law. In such cases, the prediction errors are too large and we cannot trust them in decision making. Therefore, Equation~\ref{eq:little} can be used to verify the predictions of the models and to filter the predictions that are too bad.

\begin{table}[t]
\caption{Pearson correlation coefficients between measurements in Figure~\ref{fig:little-law} for Little's equation's left and right parts.}
% \label{table}
% \small
% \begin{tabularx}{\linewidth}{X@{}X@{}X@{}X@{}X@{}}
%\begin{tabular*}{17.5pc}{@{}|p{30pt}|p{112pt}|p{30pt}|@{}}
\begin{tabular}{p{33pt}p{40pt}p{38pt}p{35pt}p{30pt}}
\hline
Sample & 
Observations & 
kNN & 
CatBoost &
NF \\
\hline
Cache     & 0.99  & 0.96    & 0.99    & 0.99 \\ 
SSD rand. & 0.99  & 0.92    & 0.97    & 0.99 \\
SSD seq.  & 0.99  & 0.89    & 0.99    & 0.99 \\
HDD seq.  & 0.91  & 0.72    & 0.92    & 0.89 \\
\hline
\end{tabular}
\label{tab:pearson}
\end{table}

In additional experiments, we measured the dependencies of the model's quality from the number of data loads used in the training sample as shown in Figures \textbf{Figure~\ref{fig:ex_hdd_seq}} and \textbf{Figure~\ref{fig:ex_cache_rnd}} for the HDD pool and cache accordingly. We used the same test sample for each measurement on the graphs. The results demonstrate that the metrics rapidly drop with increasing the number of data loads. But then, the decline in the dependencies is slowing down and they are approaching the plateau. SSD pools under random and sequential data loads have similar dependencies. They show we have enough observations in our samples to achieve reasonable quality of the performance modeling, and additional measurements will provide minor improvements.

\begin{figure*}[h!]
\centerline{\includegraphics[width=1.1\textwidth]{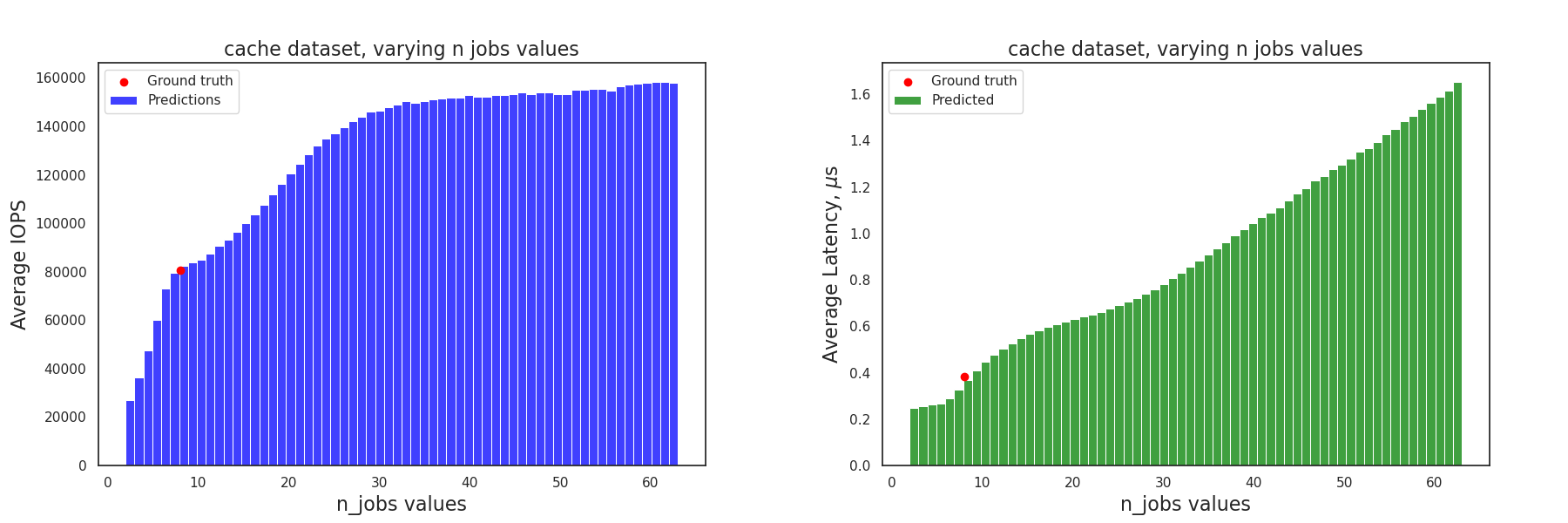}}
\caption{Exploring cache performance based on one parameter changing. Varying values for the number of parallel jobs from 2 to 64, while fixing every other parameter. Fixed values: block size = 32, iodepth = 8, read fraction = 64, load type = random, io type = write}
\label{fig:njobs_varying}
\end{figure*}

We also investigate how sensitive our models are under the input change by varying one parameter while fixing every other and then obtaining predictions. For example, in \textbf{Figure~\ref{fig:njobs_varying}} we demonstrate how our normalizing flow model predicts IOPS and latency in randomly sampled observation. We fixed every parameter, except the number of jobs: block size = 32, iodepth = 8, read fraction = 64, load type = random, io type = write. The red dot on the plots shows the ground truth system performance. In the IOPS plot, we observe a faster rate of growing IOPS on small numbers of parallel jobs, then a slow growth, and saturation at the end. We also observe the same change in the rate of growth in latency, but at a different scale.

\section{Discussion}

% \begin{figure*}[H]
% \centering
% \begin{subfigure}{1\columnwidth}
%     \centering
%     \includegraphics[width=\textwidth]{hdd_seq_anomaly_raid42_read.pdf}
%     \caption{Example of real observations in the train and test samples for the HDD pool.}
%     \label{fig:anomaly_hdd}
% \end{subfigure}%
% \hfil
% \begin{subfigure}{1\columnwidth}
%     \centering
%     \includegraphics[width=\textwidth]{hdd_seq_anomaly_hse-08302022-181607-z62969.pdf}
%     \caption{Example of real observations and CatBoost predictions for one data load on the HDD pool.}
%     \label{fig:anomaly_hdd_exam}
% \end{subfigure}

% \begin{subfigure}{1\columnwidth}
%     \centering
%     \includegraphics[width=\textwidth]{cache_anomaly_read.pdf}
%     \caption{Example of real observations for read data loads in the train and test samples for the cache.}
%     \label{fig:anomaly_cache}
% \end{subfigure}%
% \hfil
% \begin{subfigure}{1\columnwidth}
%     \centering
%     \includegraphics[width=\textwidth]{cache_anomaly_hse-09132022-215639-z27019.pdf}
%     \caption{Example of real observations and CatBoost predictions for one data load on the cache.}
%     \label{fig:anomaly_cache_exam}
% \end{subfigure}
% \caption{Anomaly Figures}
% \end{figure*}

The results in the previous section show that generative models are suitable for performance modeling tasks. They are able to simulate the performance of a data storage system and its components for the given data load and configuration of the system. The models learn the conditional distribution of IOPS and latency, which can be used to predict their average values, as well as to estimate the variance of the predictions, confidence intervals, and other useful statistics.

The results demonstrate distinct patterns in model performance based on input parameters. For instance, as shown in Figures 4 and 5, both models capture trends in IOPS and latency under varying queue depths and read/write fractions, with CatBoost showing consistently lower errors for standard deviation predictions. Additionally, the Normalizing Flow model's conservative behavior in extrapolation scenarios, such as those depicted in Figure 9, highlights its tendency to bias predictions toward the training data in rare configurations. By combining insights from these patterns, we can better understand the trade-offs between precision and distribution modeling across configurations.

Samples with high error values tend to be near the boundary of the training data space, where the model operates in an extrapolation regime. An example of this behavior in the HDD predictions is presented in \textbf{Figure \ref{fig:anomaly_hdd}}, where one test data sample is indicated by the crosshairs (dashed lines), which is zoomed in in \textbf{Figure \ref{fig:anomaly_hdd_exam}}. The predictions for this particular sample are also shown in this plot. This sample demonstrates what we call model conservatism, which is an instance of a model's predictions being biased and shifted towards the training data in case the test data sample is the extrapolation regime. This figure represents the predictions being shifted toward the test samples, which shows that the model is consistent with the data on which it was trained. 

The same pattern of behavior is demonstrated in \textbf{Figure \ref{fig:anomaly_cache}}. Here, the crosshairs are pointed on the test sample which is focused on in \textbf{Figure \ref{fig:anomaly_cache_exam}} together with the predictions. It is observed that the predictions are located close to the test samples with the latter being in the extrapolation regime. This is another example of the model conservatism.

\section{Conclusion}
Storage system simulation models are indispensable tools in the design, optimization, and management of data storage systems. These models enable a detailed analysis of performance metrics such as throughput, latency, and I/O operations per second (IOPS), providing insights into how storage systems perform under various conditions. By replicating real-world workloads, simulation models help identify bottlenecks and optimize resource allocation. They serve as crucial instruments for designing and fine-tuning storage system architectures, allowing engineers to evaluate the impact of different configurations and workload scenarios. This leads to optimized system parameters that achieve desired performance goals. Furthermore, simulation models offer a cost-effective alternative to risky and expensive live environment changes, enabling extensive testing and validation of new technologies and configurations in a controlled setting. Their predictive analysis capabilities are essential for capacity planning, ensuring that storage systems can be efficiently scaled to meet future demands. Moreover, they facilitate detailed cost-benefit analysis, helping organizations evaluate the financial implications of different storage technologies and configurations, and assess the economic viability of upgrades and expansions. In addition to performance benchmarking, simulation models play a key role in energy efficiency by optimizing the storage system to minimize energy requirements.

This work shows the results of the performance modeling of a data storage system using generative models. The outcomes help to conclude the following statements:

\begin{itemize}

\item Generative models are reasonable alternatives to other methods for performance modeling studies. This approach is suitable for single devices and their combinations.
\item Both parametric and nonparametric models demonstrate similar prediction qualities for mean IOPS and latency values. However, the parametric model estimates the standard deviations better.
\item The models can be used to predict the performance of a data storage system and its components for the given data load and configuration parameters.
\item The task has an unsupervised way for prediction reliability check based on Little's law. The models we consider in this paper demonstrate this liability.
\item We provide real measurements of IOPS and latency for the cache, SSD, and HDD pools of a data storage system. This data set can be used in future works in the field of data-driven performance modeling and studies related to conditional generative models, uncertainty estimation, and model reliability.

\end{itemize}

Scripts of all our experiments in this work and all data sets are provided in the GitHub repository\footnote{\url{https://github.com/HSE-LAMBDA/digital-twin}}.

There are multiple constraints associated with the methods presented in this work. The models need data for training, and the accuracy of the predictions relies heavily on the size of the sample. It is necessary to employ multiple system configurations and generate a diverse range of data loads. The data collection process requires access to a real system and can be time-consuming. In addition, the models show sensitivity to the coverage of the input parameter space. Unreliable predictions can occur when the training sample lacks measurements from specific regions of configurations and data load parameter values. We can not rely on predictions for the external measurement regions because of the absence of relevant information in the training data. 

There are several ways of future work for further improvement of the presented approach. The first is dedicated to increasing the liability of the models by estimating the uncertainties of the predictions. The predictions of the machine learning models are not ideal and the error values can be different for different inputs. Estimating prediction errors for each given data load parameter will help us decide whether we can rely on the prediction or not. The second direction is related to the explainability of the models. Currently, the models make performance predictions and do not provide an understanding of how the inputs affect the result. Information about how the prediction changes with changing the read fraction, queue depth, number of jobs, or other parameters will give a better understanding of the behavior of the system and will help optimize its performance.

\bibliographystyle{IEEEtran}
\bibliography{references.bib}

\begin{IEEEbiography}[{\includegraphics[width=1in,height=1.25in,clip,keepaspectratio]{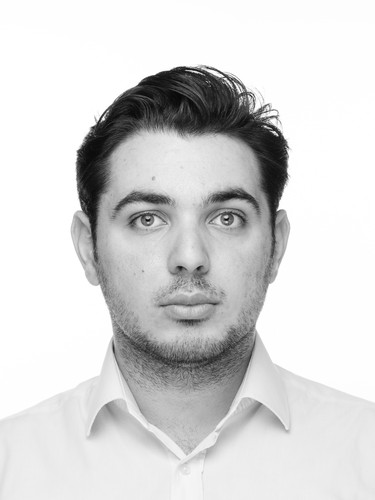}}]{Abdalaziz R. Al-Maeeni} 
Abdalaziz R. Al-Maeeni received his M.Sc. degree from the National University of Science and Technology (MISiS). He is currently pursuing his Ph.D. in computer science and serving as a Junior Research Fellow at the Faculty of Computer Science, HSE University in Russia. His academic focus is primarily on the development and analysis of interpretable generative models and their applications in inverse design. His work aims to enhance the transparency and usability of generative models in designing systems and solutions across various domains.
\end{IEEEbiography}

\vskip -2\baselineskip plus -1fil % adjust this line depending on the biographies size

\begin{IEEEbiography}[{\includegraphics[width=1in,height=1.25in,clip,keepaspectratio]{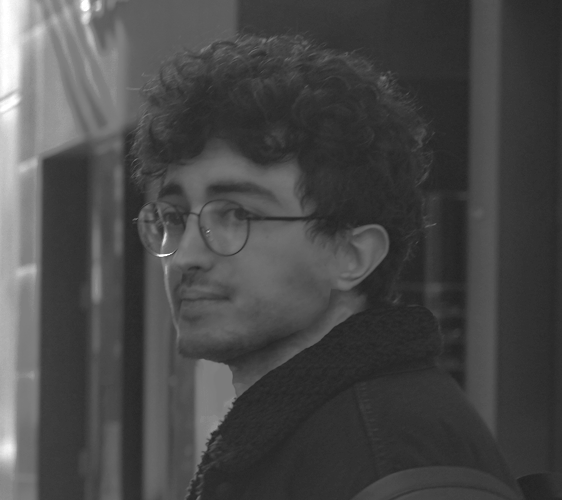}}]{Aziz Temirkhanov} is a PhD student at the Faculty of Computer Science, HSE University, Russia. His main research interests are generative models and its application in natural science.
\end{IEEEbiography}

\vskip -2\baselineskip plus -1fil % adjust this line depending on the biographies size

\begin{IEEEbiography}[{\includegraphics[width=1in,height=1.25in,clip,keepaspectratio]{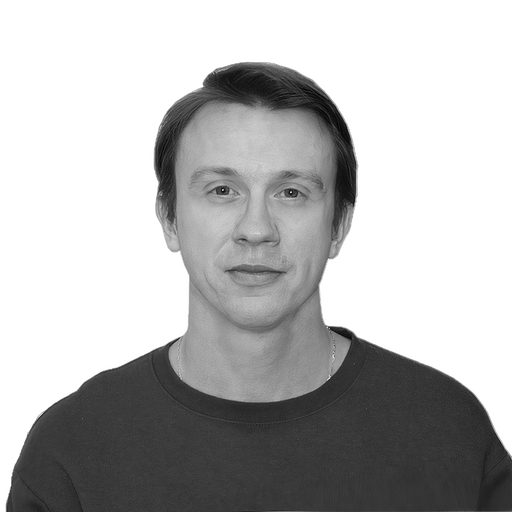}}]{Artem Ryzhikov} received the M.Sc. degree and getting PhD in Computer Science from National Research University Higher School of Economics, Russia. His research interests are in the area of machine learning and its application to High-Energy Physics with a focus on Deep Learning and Bayesian methods. 
\end{IEEEbiography}

\begin{IEEEbiography}[{\includegraphics[width=1in,height=1.25in,clip,keepaspectratio]{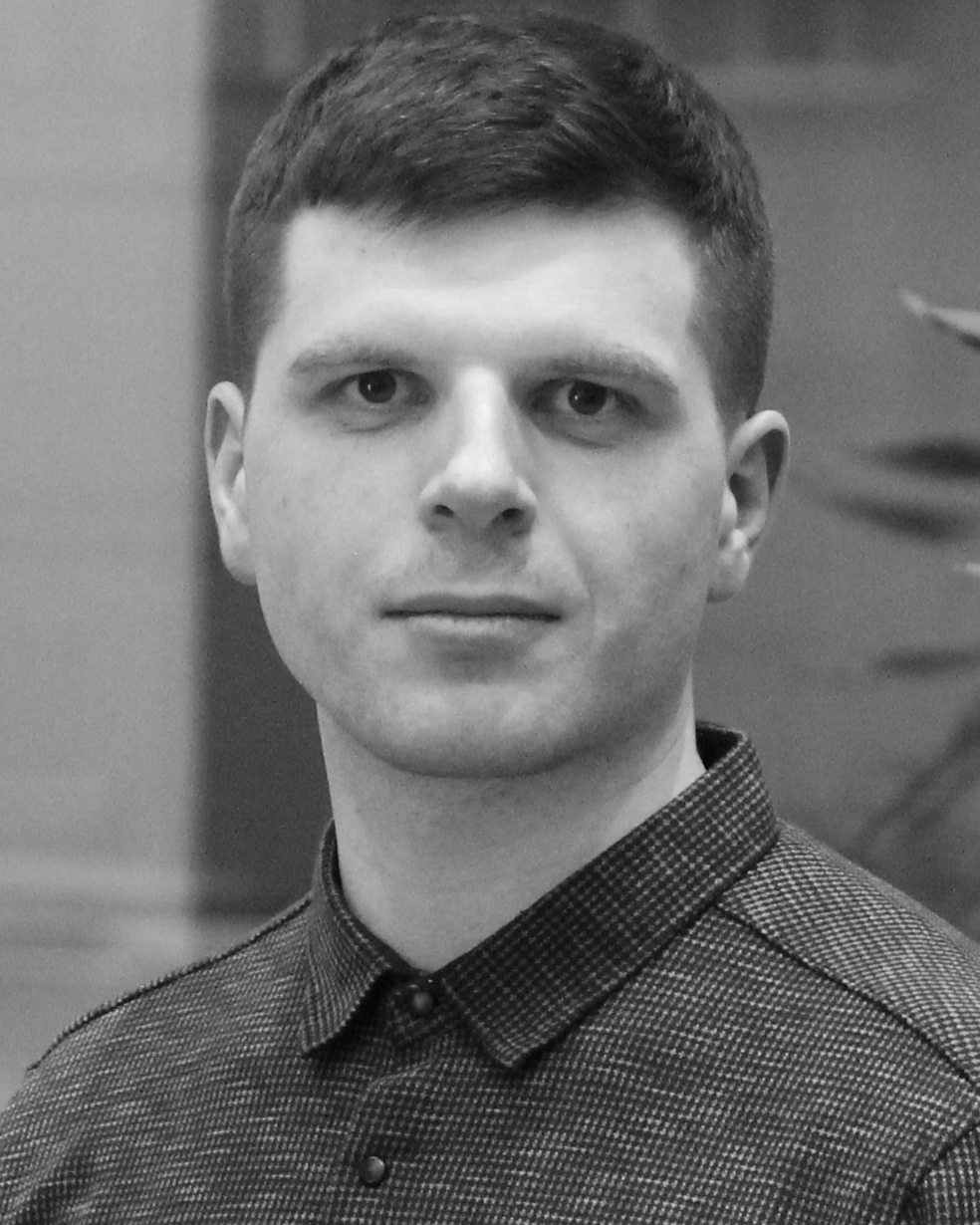}}]{Mikhail Hushchyn} received the MSc degree in 2015 in the Moscow Institute of Physics and Technology (MIPT) and obtained in 2019 a PhD in mathematical modeling, numerical methods and program complexes from the same university. He joined HSE University, Moscow and currently is a Senior Research Fellow there. His main research interests include the application of machine learning and artificial intelligence methods in the natural sciences and industry.
\end{IEEEbiography}

\EOD

\end{document}